# **Reinforcement Learning in Medical Image Analysis:**
# Concepts, Applications, Challenges, and Future Directions

Mingzhe Hu[a], Jiahan Zhang[b], Luke Matkovic[b], Tian Liu[b] and Xiaofeng Yang[a,b*]

[a]*Department of Computer Science and Informatics, Emory University, GA, Atlanta, USA*
[b]*Department of Radiation Oncology, Winship Cancer Institute, School of Medicine, Emory University, GA, Atlanta, USA*

**Email: Xiaofeng.yang@emory.edu*

**Abstract**

**Motivation**: Medical image analysis involves tasks to assist physicians in qualitative and quantitative analysis of lesions or anatomical structures, significantly improving the accuracy and reliability of diagnosis and prognosis. Traditionally, these tasks are finished by physicians or medical physicists and lead to two major problems: (i) low efficiency; (ii) biased by personal experience. In the past decade, many machine learning methods have been applied to accelerate and automate the image analysis process. Compared to the enormous deployments of supervised and unsupervised learning models, attempts to use reinforcement learning in medical image analysis are scarce. This review article could serve as the stepping-stone for related research.
**Significance**: From our observation, though reinforcement learning has gradually gained momentum in recent years, many researchers in the medical analysis field find it hard to understand and deploy in clinics. One cause is lacking well-organized review articles targeting readers lacking professional computer science backgrounds. Rather than providing a comprehensive list of all reinforcement learning models in medical image analysis, this paper may help the readers to learn how to formulate and solve their medical image analysis research as reinforcement learning problems.
**Approach & Results**: We selected published articles from Google Scholar and PubMed. Considering the scarcity of related articles, we also included some outstanding newest preprints. The papers are carefully reviewed and categorized according to the type of image analysis task. We first review the basic concepts and popular models of reinforcement learning. Then we explore the applications of reinforcement learning models in landmark detection. Finally, we conclude the article by discussing the reviewed reinforcement learning approaches' limitations and possible improvements.

## 1. Introduction

The purpose of medical image analysis is to mine and analyze valuable information from medical images by using digital image processing to assist doctors in making more accurate and reliable diagnoses and prognoses. According to different imaging principles, common imaging modalities can be categorized as CT, MR, Ultrasound, SPECT, PET, X-ray, OCT, and microscope. Medical image processing can also be classified according to specific processing tasks. Typical tasks include classification, segmentation, registration, and recognition. Figure 1 shows the range of our review article.

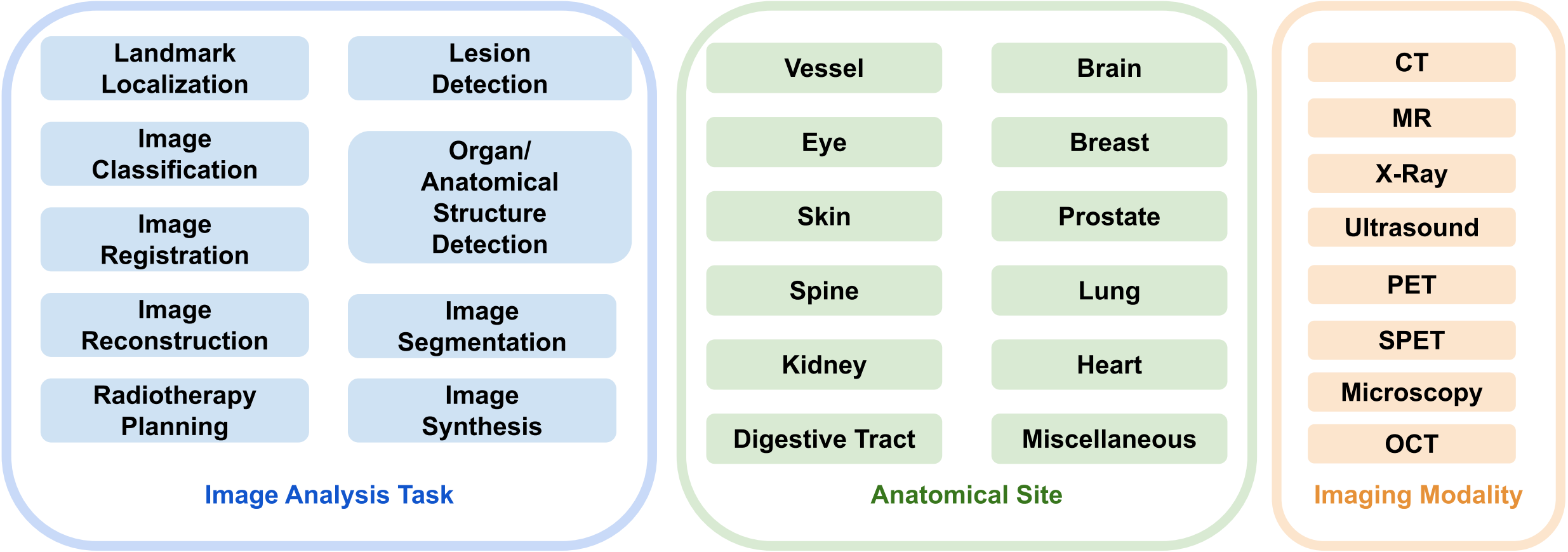

**Figure 1**: Range of our review article. Blue box: covered image analysis tasks; green box: covered anatomical sites; yellow box: covered imaging modalities.

With the development of imaging technology and the iterative update of imaging equipment, the time required for medical imaging is greatly shortened, and the resolution of imaging is also significantly improved. At the same time, the data volume of medical images has experienced an unprecedented surge, with the trend of high dimensionality. The traditional manual analysis of medical images by physicians became tedious and inefficient. More and more physicians are looking to automate this process by partnering with engineers. That's how the combined medical imaging and machine learning field was born. Many excellent algorithms in the field of natural image analysis have also shown good results in the field of medical images (Shen et al., 2017).

Reinforcement learning (RL) is neither supervised learning nor unsupervised learning. The goal of reinforcement learning is to achieve the maximum expected cumulative reward (Sutton & Barto, 2018).

Figure 2 shows the relationship between machine learning, supervised learning, unsupervised learning, reinforcement learning and deep learning.

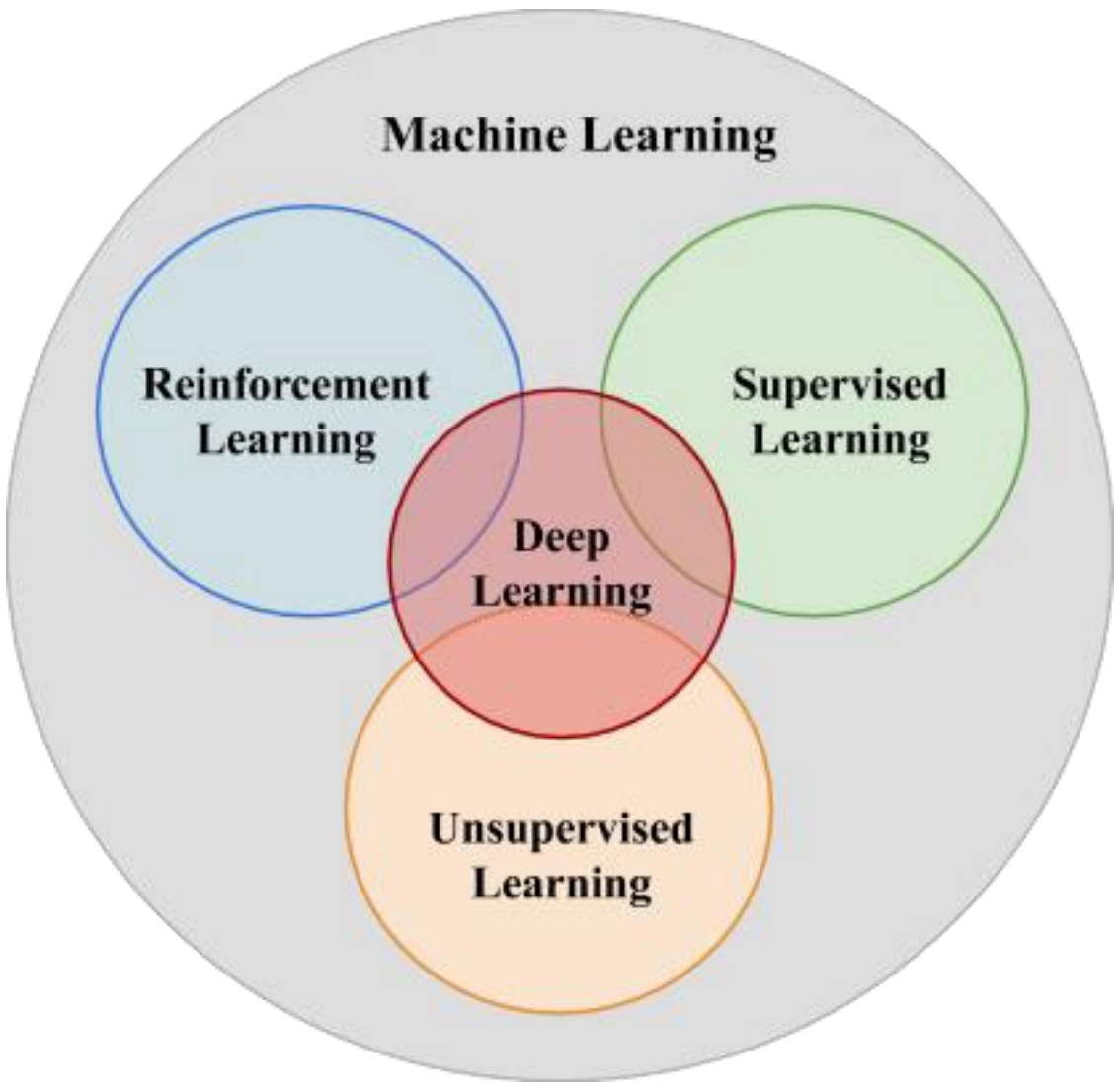

**Figure 2**: Relationship between machine learning, supervised learning, unsupervised learning and reinforcement learning.

The number of published reinforcement learning-related papers has grown rapidly in the past two decades. State-of-the-art RL models have been applied to solve problems that are difficult or infeasible with other machine learning approaches, such as playing video games (Mnih et al., 2013; Mnih et al., 2015; Silver et al., 2017), natural language processing (Sharma & Kaushik), and autonomous driving (Sallab et al., 2017). These RL methods have achieved outstanding performances. However, attempts to exploit the technical developments in RL in the medical analysis field are scarce. Figure 3 shows the trends of number of published machine learning papers and reinforcement learning papers in medical image analysis. Despite the overall growth trend, the number of published RL papers still only constitutes a tiny part of machine learning in medical image analysis. On the other hand, RL methods have unique advantages in dealing with medical image data:

- RL models can efficiently learn from limited annotation guided by supervised actions step by step, while medical data often lacks large-scale accessible annotation.
- RL models are less biased since they won't inherit bias from the labels made by human annotators.
- RL-agents can learn from sequential data, and the learning process is goal-oriented. Besides exploiting experience, it can also explore new solutions. The RL can even surpass human experts when solving the same problem.

The review article is based on Synthesis Methodology (Wilson & Anagnostopoulos, 2021).

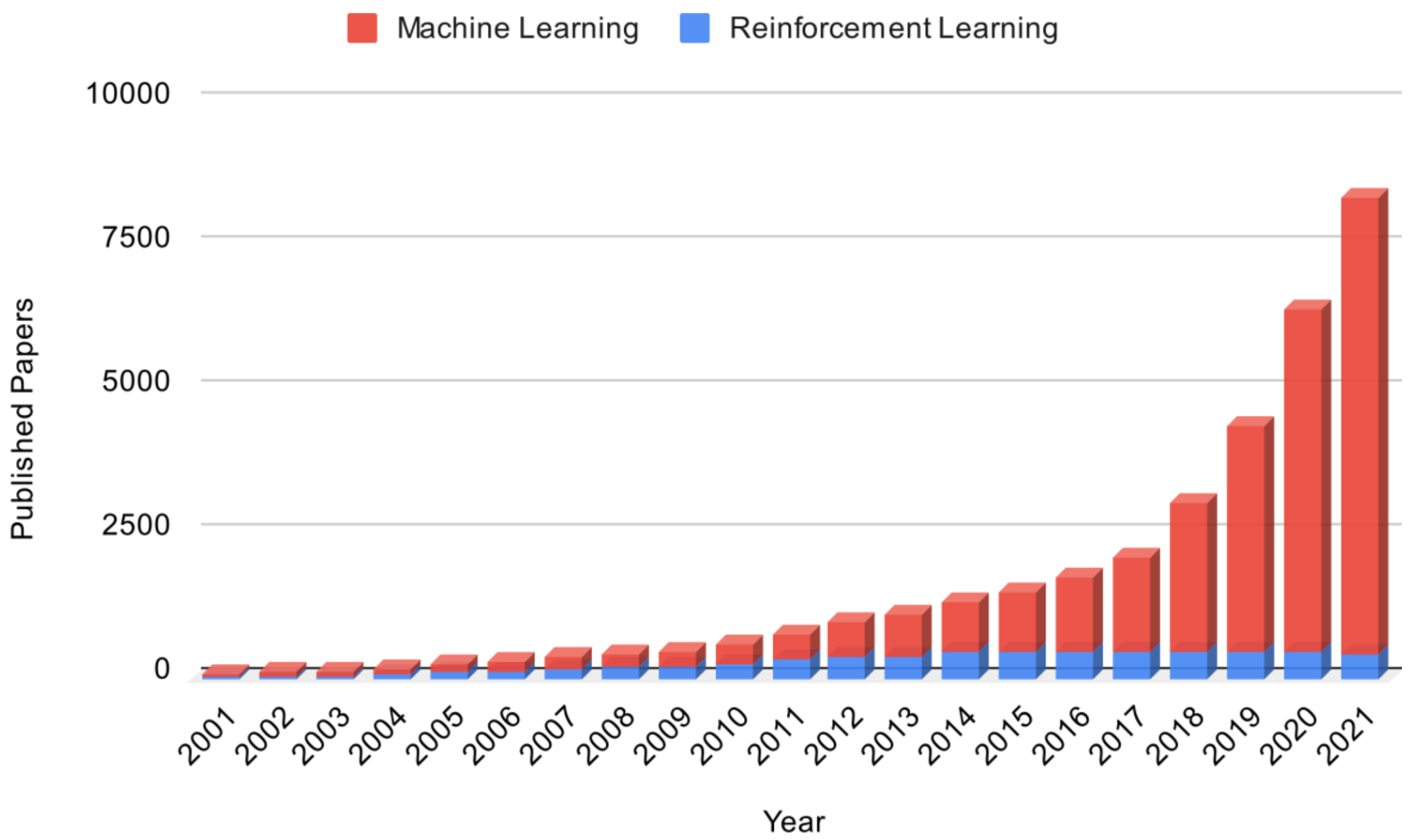

**Figure 3**: Trends of number of published machine learning papers and reinforcement learning papers in medical image analysis. This figure is made by separately searching the keywords "Machine Learning AND (Medical Imaging OR (Medical Image Analysis))" and "Reinforcement learning AND (Medical Imaging OR (Medical Image Analysis))" in PubMed. The number of papers published each year is counted.

Preferred Reporting Items for Systematic Reviews and Meta-Analyses (PRISMA) will be followed(Moher et al., 2009). Firstly, the following pattern will be searched in Google Scholar and PubMed: Clustering AND (Medical OR CT OR MR OR Ultrasound OR X-ray OR OCT) AND IMAGE AND Segmentation. Then the duplicate papers will be removed. We set the qualified publication date to 2010. The remaining papers will go through qualitative synthesis and quantitative synthesis. The summary of the selection process is shown in Figure 4.

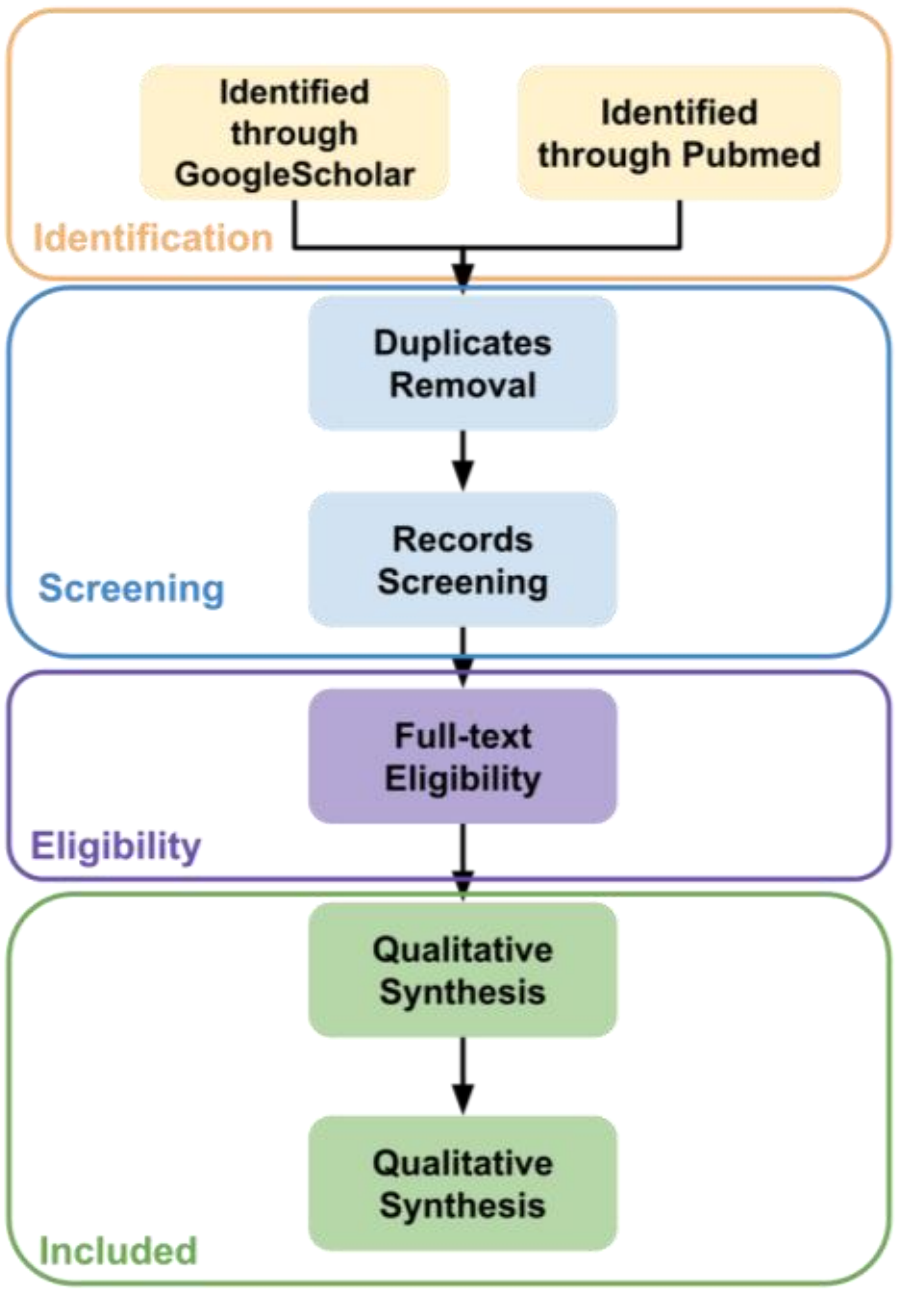

**Figure 4**: Flows of information through the different phases of a systematic review.

By reviewing content, analyzing common points and comparing difference of these papers, we hope that we can inspire our target readers to (i) have a better understanding of RF, (ii) learn how to formulate their research problems as RL problems. For the next two sections, we will first prepare the readers with basic knowledge of RL. Then we will show how to apply RL in different medical image analysis tasks. Those readers who have alreadybeen familiar with RL algorithms could directly go to the application section.

## 2. Reinforcement Learning Basics

In this subsection, we provide a list of terminologies that frequently appear in RL papers. Some terminologies may appear in definitions of other terminologies before they are defined.

- *Action (A)*: An action (a) is the way that an agent interacts with the environment. A includes all possible actions that an agent could perform.
- *Agent*: Agents are the models we attempt to build that interact with the environment and take actions.
- *Environment*: The content that the agent is interacting with is called the environment. While providing feedback after the agent takes action, the environment itself is also changing.
- *State (S)*: A state (s) is a frame of an environment. S includes all states that an agent will go through.
- *Reward (R)*: A positive reward (r) means an increase possibility of achieving the goal, while a negative reward means the decreased possibility. R includes all the possible reward values the environment may feed back to the agent.
- *Episode*: If an agent has gone through all the states from the initial state to the terminal state, we say this agent has finished the episode.
- *Transition probability* $P(s'|s,a)$: $P(s'|s,a)$ is the possibility of transiting to transiting to state $s'$ to from the current state s, taking the action a.
- *Policy* $\pi(a|s)$: The policy instructs the agent to choose among actions A under the current state.
- *Return (G)*: The return is the cumulative discounted future reward.
- $G_t = r_t + \gamma r_{t+1} + \gamma^2 r_{t+2}$, where t is the time and γ is the discount factor.
- *State value* $V^{\pi}(s)$: The expected amount of return from current state.
- $V^{\pi}(s) = E[G_t | s_t = s]$, where E is the expectation.
- *Action value* $Q^{\pi}(s,a)$ *(Q value)*: The expected amount of return from current state, taking action s. $Q^{\pi}(s,a) = E[G_t | s_t = s, a_t = a]$
- *Optimal action value*: $Q^{\star}(s,a)$: $Q^{\star}(s, a) = \max_{\pi} Q^{\pi}(s_t, a_t)$
- *Agent environment interaction*: Figure 5 shows how the agent is interaction with the environment.

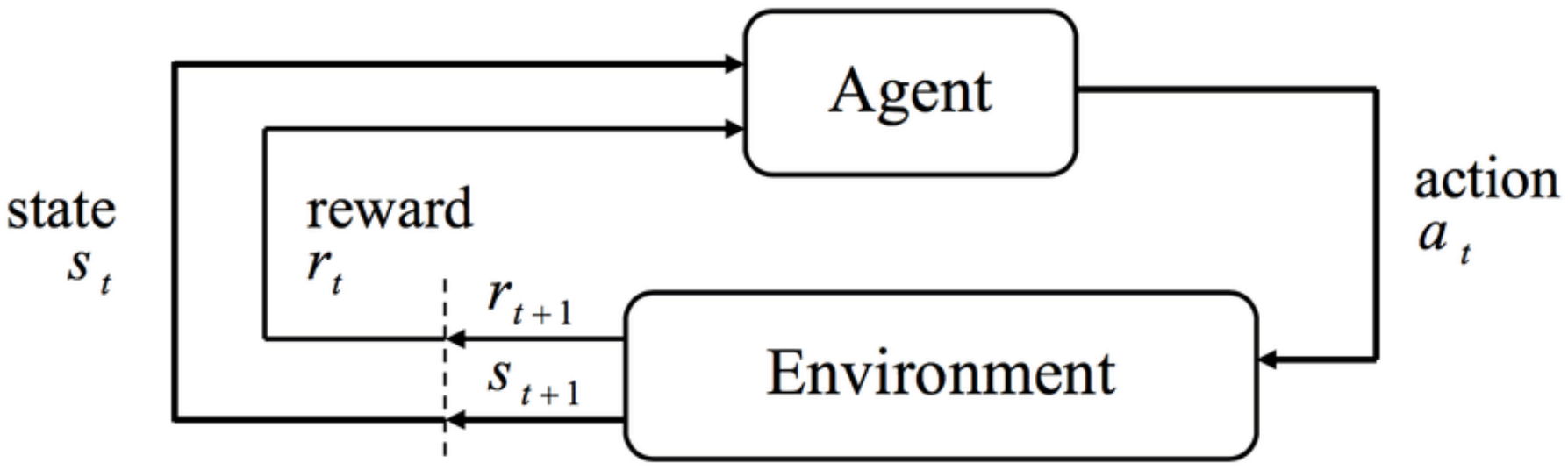

**Figure 5**: Agent Environment Interaction. Adapted from (Rafati & Noelle, 2019).

With the development of the RL theory, numerous algorithms have been created. Benefiting from the combination with deep learning, RL is now capable of handling more and more complex scenarios in modern applications. But no matter how complex these state-of-the-art algorithms are, they can be mainly divided into two categories: model- based RL and model-free RL. As its name indicates, model-based RL attempts to explain the environment and create a model to simulate it. Model-free RL, however, will only update its policy by interacting with the environment and observing the rewards.

We can further divide the model-free RLs into policy-based and value-based according to whether the algorithm is optimizing the value function or policy. Value-based RLs are widely applied for discrete action space problems, while policy-based RLs are suitable for both discrete and continuous action space. Some RL algorithms are based on both the value and policy, like DDPG (Lillicrap et al., 2015), TD3 (Fujimoto et al., 2018) and SAC (Haarnoja et al., 2018). Figure 6 shows the taxonomy of popular RL algorithms. In our review, all the RL models are model-free, and the mostly used algorithms are DQN, DDQN, A2C, and DDPG. Below we include brief introductions of these RL algorithms commonly used in medical image analysis.

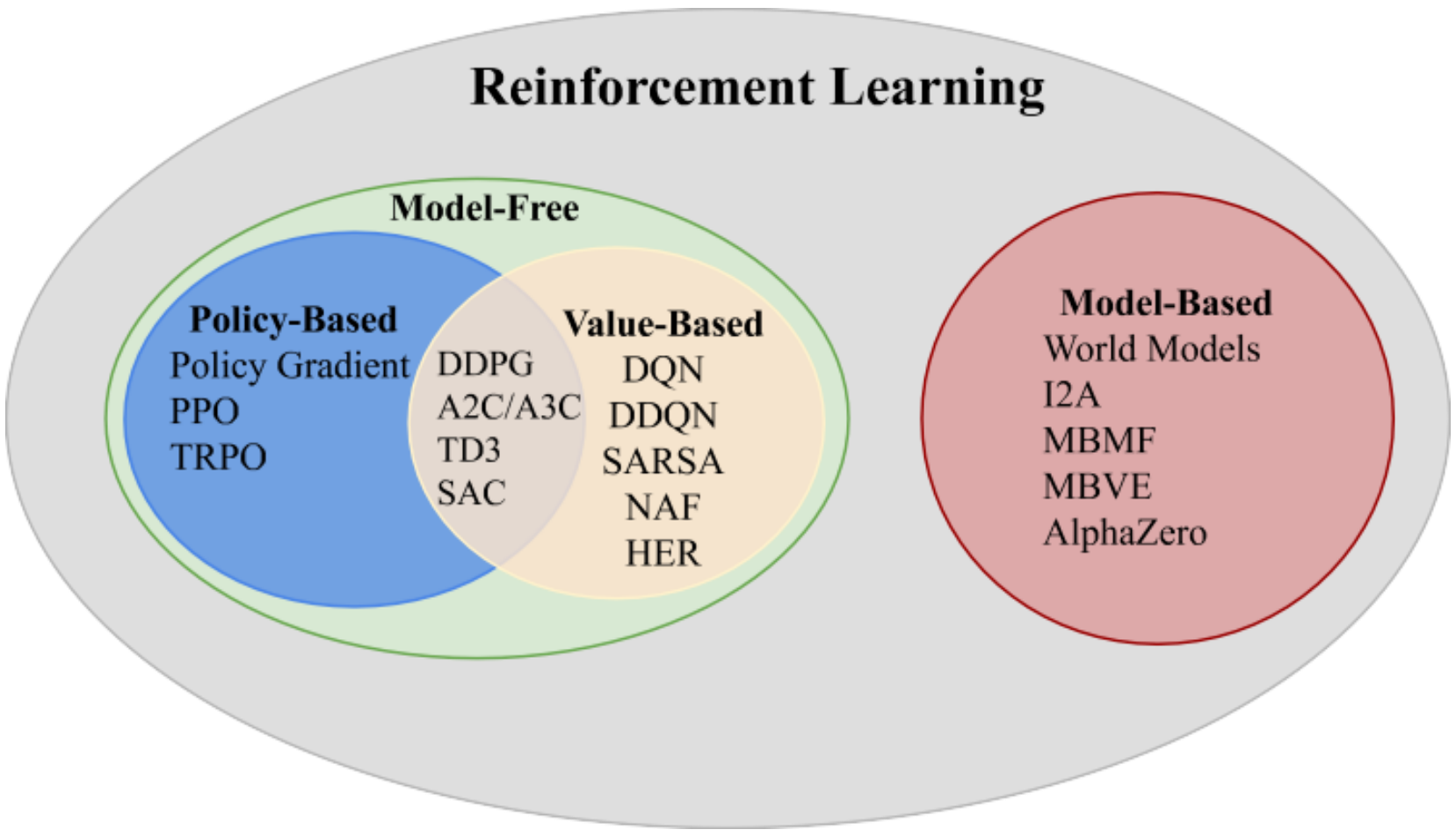

**Figure 6:** Reinforcement learning algorithms taxonomy.

## *2.1. DQN*

The Deep Q-Network (DQN) was first proposed by (Mnih et al., 2013; Mnih et al., 2015) to solve some complex computer perception vision problems. It combined the idea of the traditional Q learning method (Watkins, 1989) and the deep CNN (Krizhevsky et al.). The motivation of DQN is to solve the problem that the Q-table can only store a limited number of states, while in real-life scenarios, there could be an immense or even infinite number of states. DQN adopts the experience replay mechanism that randomly samples a small batch of tuples from the replay buffers during the training process. The correlations between the samples are significantly reduced, leading to better algorithm robustness. Another improvement, compared to Q learning, is that DQN uses a deep CNN to represent the current Q function and uses another network to define the target Q value. The introduction of the target Q value network reduced the correlation between the current and target Q values. Figure 7 shows the workflow of the DQN.

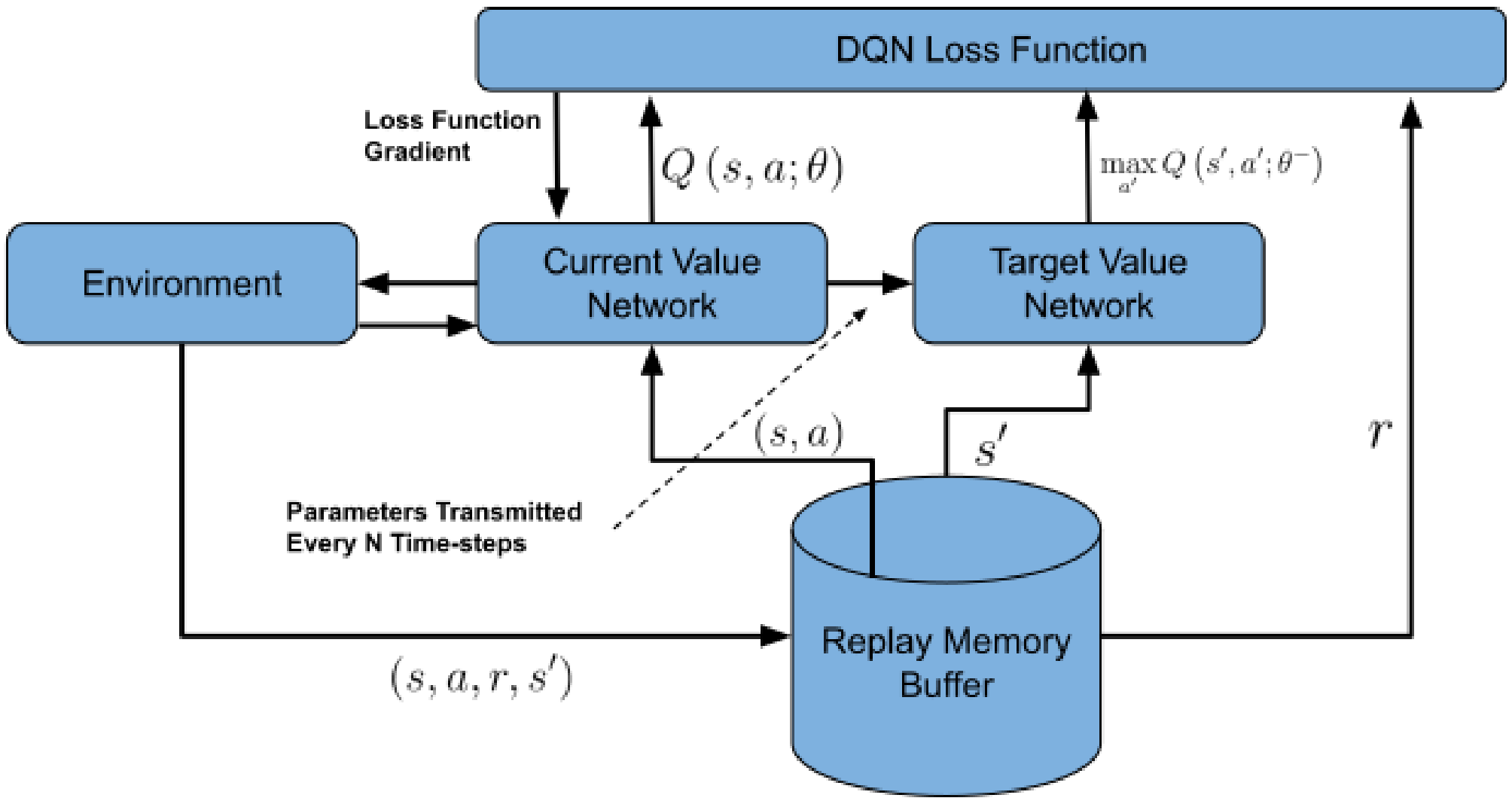

**Figure 7**: Workflow of DQN algorithm.

## 2.2. *DDQN*

DQN is one of the most popular RL algorithms applied in medical image analysis. How- ever, the optimization target in DQN is represented as $r + \gamma \max_{a'} Q(s', a'|\theta_i^-)$. The selection and evaluation of actions are all based on the network's same parameter, leading to over- estimation of the Q value. The Double DQN (DDQN), which (Van Hasselt et al.) first proposed, used two separate networks for selection and evaluation. Here the target Q value is written as $r + \gamma Q(s', \text{argmax}_a Q(s', a|\theta_i), \theta_i^-)$, which achieved better more stable learned policy than DQN.

## 2.3. *Actor Critic*

The Actor-Critic (AC) (Konda & Tsitsiklis, 1999) algorithm mixes the idea of policy gradient and Time Difference (TD) learning and can handle continuous action space problems and update the policy in an efficient stepwise manner. The actor is the policy function $\pi_\theta(a|s)$ that learns the policy using gradient descent to achieve the highest possible reward. And the critic is the value function $V^\pi(s)$ that uses the TD error to assess the current policy.

## 2.4. *A2C/A3C*

A2C (Sewak, 2019) and A3C (Mnih et al., 2016) are improved versions of the vanilla actor-critic that introduced the parallel architecture. Each agent includes a global network and multiple workers that run independently. Every worker would gather different experiences and calculate the different gradients. According to the ways of tackling the different parameters and different gradients, we can derive the synchronous version — A2C and asynchronous version — A3C. Synchronous here means that different workers share the same policy, and the time to update the policy is the same, making the A2C tend to converge faster than the A3C.

## 2.5. *DDPG*

DDPG is a type of actor-critic-based algorithm but learns the off-policy. Similar to DQN, samples generated by the random policy are stored in the memory replay buffer. However, DQN can only solve control problems with discrete actions, while DDPG can solve the problems in the continuous action space and shows excellent efficiency in finding the optimal policy. However, for some random environments, such as low signal-to-noise-ratio images, the deterministic policy gradient strategy adopted by DDPG is not suitable.

# 3. RL in Medical Image Analysis

## 3.1. *Medical Image Detection*

### 3.1.1. *Overview of Works*

*Landmark Detection*

Anatomical landmarks are biological coordinates that can be reallocated repeatedly and precisely on images produced by different imaging modalities — computed tomography (CT), ultrasound (US), magnetic resonance imaging (MRI). The accurate detection of anatomical landmarks is the ground for further medical image analysis tasks. Figure 8 is an example of vocal tract landmarks from the MRI image.

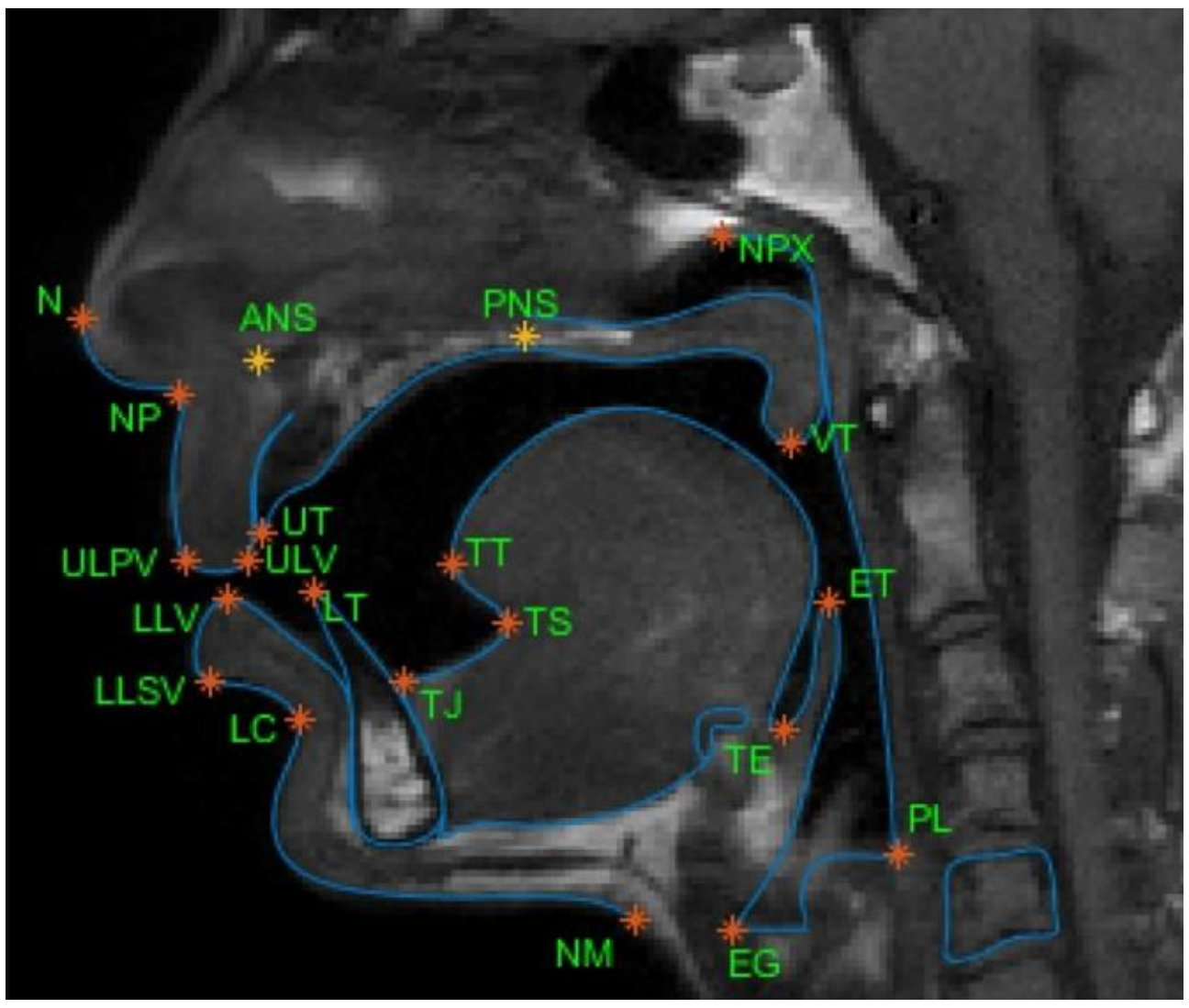

**Figure 8**: Vocal tract landmarks from MRI image Courtesy of (Eslami et al., 2020).

Many automatic algorithms for anatomical landmark detection have existed long before the attempts of using RL models. However, landmark detection, especially 3D landmarks detection, could be challenging and cause the failure of these algorithms (Ghesu et al., 2019). Moreover, the computation of features and hyper-parameters selection of the system may not be optimal since the involvement of human decisions. The researchers attempted a different paradigm to address this problem - translate the landmark detection tasks as reinforcement learning problems which is the common goal of the papers we reviewed. While most essential and tricky task in these papers, as you can see later, is designing the state space, action space, and reward space before training the models.

(Ghesu et al., 2016) is one of the very first papers that attempted to use RL for anatomical landmark detection. In an image I, $\overrightarrow{p_{GT}}$ denotes the location of anatomical landmark, and $\overrightarrow{p_t}$ denotes the location at the current time. State space S is the collection of all possible states $s_t = I(\overrightarrow{p_t})$. Action space A is the collection of all possible actions by which the agent can move to the adjacent position, as illustrated by Figure 9. Reward space R is defined as $||\overrightarrow{p_t} - \overrightarrow{p_{GT}}||_2^2 - ||\overrightarrow{p_{t+1}} - \overrightarrow{p_{GT}}||_2^2$ which impels the agent to move closer to the target anatomical landmark. A deep learning model was applied to approximate the state value

function. The parameters are updated according to gradient descent, and the error function is:

$$\widehat{\theta_i} = \arg\min_{\theta_i} E_{s,a,r,s'}\left[\left(y - Q(s,a;\theta_i)\right)^2\right] + E_{s,a,r}[V_{s'}(y)] \tag{1}$$

This deep Q learning-based method beat the existing top systems not only in accuracy but also in speed. The design of action, state, and reward spaces in the paper we just discussed became a standard method.

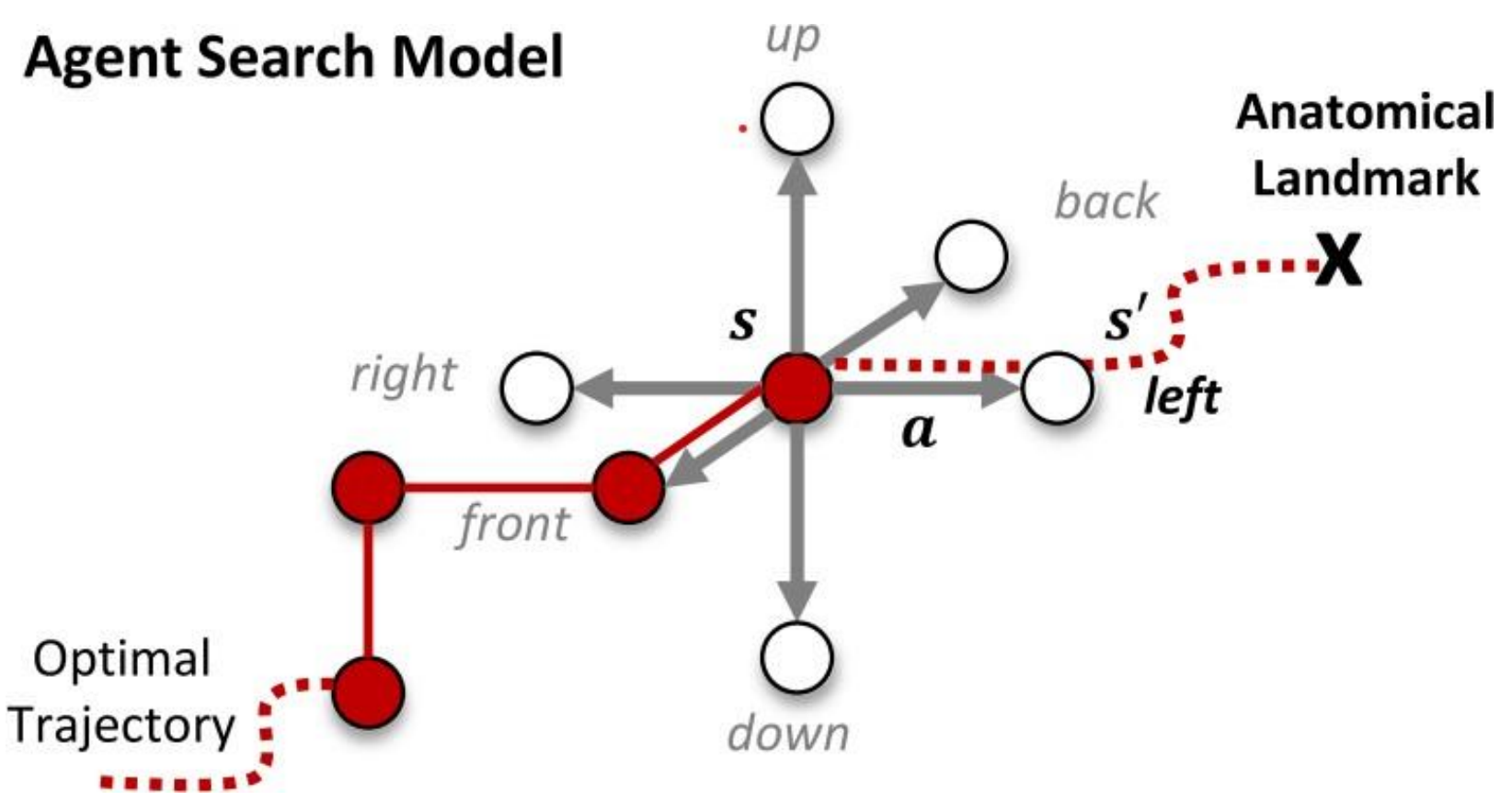

**Figure 9**: Possible actions of a 3D landmarks detection task. Courtesy of (Ghesu et al., 2019).

However, the approach mentioned above is still preliminary. One of the biggest disadvantages is that it could not fully use the information at different scale. So a multi-scale deep reinforcement learning method was soon proposed in (Ghesu et al., 2019). The search for the landmark started from the coarsest scale. Once the search is convergent, the continued work would be started at a finer scale until the search meets the finest scale's convergence criteria. Figure 10 illustrates this non-trivial search process. Where $L_d$ is the scale level in the continuous scale-space L, which can be calculated as:

$$L_d(t) = \psi_\rho\left(\sigma(t-1) * L_d(t-1)\right) \tag{2}$$

Where $\psi_\rho$ is the signal operator, and $\sigma$ is the Gaussian-like smoothing function.

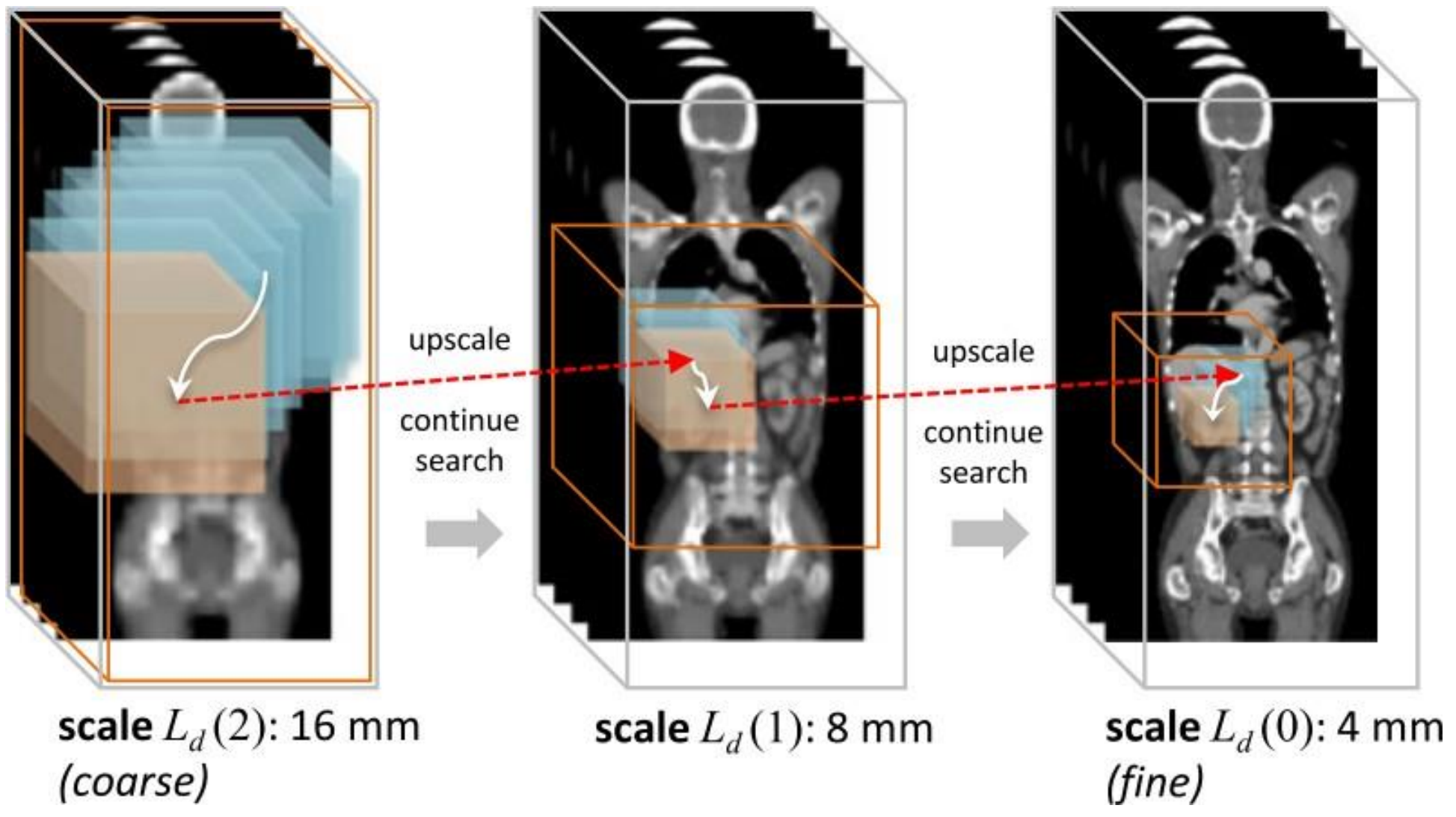

**Figure 10**: The trajectory of search the anatomical landmark across images of multiple scale-levels. Courtesy of (Ghesu et al., 2019).

(Alansary et al., 2019) extended the work of Ghesu et al. by evaluating different types of RL agents. He compared the detection results of using DQN, double DQN (DDQN), Double DQN, and duel double DQN (duel DDQN) on three different-modalities dataset — fetal US, cardiac MRI, and brain MRI.

Rather than detecting a landmark per agent separately, bold attempts have been made by (Vlontzos et al., 2019) to detect multiple landmarks with multiple collaboration agents. With the assumption that the anatomical landmarks have inner correlations with each other, the detection of one landmark could indicate the location of some other landmarks. For the action function approximator in this paper, the collaborative deep Q network (Collab-DQN) was proposed. The weights of the convolutional layers are shared by all the agents, while the fully connected layers for deciding the actions are trained separately per agent. Compared to the methods that trained agents for different landmarks differently, this multi-agent approach reduced 50% detection error using a shorter training time.

Some other contributions to the RL for anatomical landmarks detection include: estimating the uncertainty of reinforcement learning agent (Browning et al., 2021), reducing the needed time to reach the landmark by using a continuous action space (Kasseroller et al., 2021), localization of modality invariant landmark (Winkel et al., 2020).

*Lesion Detection*

Object detection, also called object extraction, is the process of finding out the class labels and locations of target objects in images or videos. It is one of the primary tasks in medical image analysis (Li et al., 2019). An exemplary detection result can be used as the basis to improve the performance of further tasks like segmentation.

The mainstream approaches for lesion detection nowadays still rely on exhaustive search methods that cost a lot of time and deep learning methods that require a large amount of labeled data. Facing the current challenges and inspired by similar problems in landmarks detection (Ghesu et al., 2016), (Maicas et al., 2017) implemented a deep Q-network (DQN) agent for active breast lesion detection. The states are defined as current bounding box volumes of the 3D DCE-MR images. The reinforcement learning agent could gradually learn the policy to choose among actions to transit, scale the bounding box, and finally localize the breast lesion. Specifically, the action set consists of 9 actions that can translate the bounding box forward or backward along the x, y, z-axis, scale up or scale down the bounding box, and trigger the terminal state. To further evaluate the effectiveness of applying reinforcement learning on lesion detection with limited data, using DQN as the agent to localize brain tumors with very small training data was attempted by (Stember & Shalu, 2020), (Stember & Shalu, 2021a).Different from (Maicas et al., 2017), the brain MR data are 2D slices. The environment is defined as the 2D slices overlaid with gaze plots viewed by the radiologist. Instead of using the bounding box, the states are the gaze plots the agent located. Three actions - moving anterograde, not moving, moving retrograde would help the agent transfer to the next state. If the agent moves toward the lesion, it will receive a positive reward, otherwise a negative one. If the agent stays still, it will receive a relatively large positive reward within the lesion area or a rather large penalization otherwise. The experiment results showed that reinforcement learning models could work as robust lesion detectors with limited training data, reduce time consumption, and provide some interpretability.

Also addressing the lack of labeled training data, (Pesce et al., 2019) exploited visual attention mechanisms to learn from a combination of weakly labeled images (only class label) and a limited number of fully annotated X-ray images. This paper proposed convolutional networks with attention feedback (CONAF) architecture and a recurrent attention model with annotation feedback (RAMAF) architecture. The RAMAF model can only observe one part of the image, which is defined as a state at a glimpse. The reinforcement learning agent needs to learn the policy to take a sequence of glimpses and finally locate the lesion site within the shortest time. Each glimpse consists of two image patches sharing the same central point, and the length of the glimpse sequence is fixed to be 7. The rewards will be decided according to (i) if the image is classified correctly; (ii) if the central point of a glimpse is within the labeled bounding box. RAMAF achieved a localization performance of detecting 82% of overall bounding boxes with a much faster detection speed than other state-of-the-art methods.

More than detecting lesions in static medical images (2D or 3D), the reinforcement- learning-based system can also track the lesions frame by frame continuously. (Luo et al., 2019) proposed a robust RL-based framework to detect and track plaque in Intravascular Optical Coherence Tomography (IVOCT) images. Despite the pollution problem of speckle- noise, blurred plaque edges, and diverse intravascular morphology, the proposed method achieved accurate tracking and has strong expansibility.

Three different modules are included in the proposed framework. The features are extracted and encoded first by the encoding feature module. Then the information of scale and location of the lesion is provided by the localization and identification module. Another function of this module is preventing over-tracking. The most important module is the spatial- temporal correlation RL module. Nine different actions are different, including eight transformation actions and one stop action. The state S is defined as three-tuples: $S = (E, HL, HA)$. Here, the E represents the encoded output features from the FC1 layer. HL is the collection of recent locations and scales. HA represents the recent ten sets of actions. 8000 IVOCT images were used to evaluate the framework. With a strict standard (IOU > 0.9), the RL module could improve the performance of plaque tracking both frame-level and plaque-level.

*Organ/ Anatomical Structure Detection*

Besides detecting lesions, reinforcement learning can also be applied in organ detection. (Navarro et al., 2020) designed a deep Q-learning agent to locate various organs in 3D CT scans. The state is defined as voxel values within the current 3D bounding box. Eleven actions, including six translation actions, two zooming actions, and three scaling actions, make sure that the bounding box can move to any part of the 3D scan. The agent is rewarded if an action improves the intersection over union score (IOU). Seventy scans were used for training, and 20 scans were used for testing on seven different organs: pancreas, spleen, liver, lung (left and right), and kidney (left and right). This proposed method achieved a much faster speed than the region proposal and the exhaustive search methods and led to an overall IOU score of 0.63.

(Zhang et al., 2021) managed to detect and segment the vertebral body (VB) simultaneously. The sequence correlation of the VB is learned by a soft actor-critic (SAC) RL agent to reduce the background interference. The proposed framework consists of three modules: Sequential Conditional Reinforcement Learning network (SCRL), FC- ResNet, and Y-net. The SCRL learns the correlation and gives the attention region. The FC-ResNet extracts the low-level and high-level features to determine a more precise bounding box according to the attention region. At the same time, the segmentation result is provided by the Y-net. The state of the RL agent is determined by a combination of the image patch, feature map, and region mask. And the reward is designed according to the change of attention-focusing accuracy to elicit the agent to achieve a better detection performance. This proposed approach accomplished an average of 92.3% IOU on VB detection and an average of 91.4% Dice on VB segmentation.

The research of (Zheng et al., 2021) was the first attempt to use the multi-agent RL in prostate detection. Two DQN agents locate the lower-left and upper-right corners of the bounding box while sharing knowledge according to the communication protocol (Foerster et al., 2016). The final location of the prostate is searched with a coarse-to-fine strategy to speed up the search process and improve the detection accuracy. In more detail, the agents first search on the coarsest scale to draw a big bounding box and gradually move to a finer scale to generate a smaller and more accurate bounding box to detect the prostate. Compared to the single-agent strategy (63.15%), this multi-agent framework achieved a better average score of 80.07% in IOU.

### 3.1.2. *Assessment*

Detection is a type of problem that straightforwardly can be formulated as the control or path-finding problem. Generally speaking, the states are defined as the pixel values that the agents observe at the current step, and the actions are defined as movements along the different axis of the environment plus some scaling factors. That is why agent-based detection has the most considerable number of papers among all RL-related image detection tasks. Though related work in this field is still growing, some challenges exist.

Firstly, the generalizability and reproducibility of the agent-based methods still need to be further investigated. In practical application, the quality and local features of the image may vary by the noise and distortion introduced in the imaging process. The trained agent may not always be capable of finding the target in clinical settings. Furthermore, the trigger of the termination state in the inference stage needs to be improved. The most commonly used criteria adopted now is the happening of oscillation. However, this may lead to a very ineffective convergence, and the agent might even be trapped at some local optimal point and never reach the actual destination. Real-time detection is another direction that has caused more interest in recent years. RL has proved its fast detection capability due to the non-exhaustive searching strategy. However, in some high dimensional data, 4D images (3D plus temporal), for example, the real-time detection and tracking still need more investigation. The last point is that the training process of the RL system, especially the multi-agent system, is very time-consuming, which may take days to weeks to train on even the best hardware platforms, let along the hyperparameter-tunning is also highly relied on the designer's experience. A summary of the works we reviewed in this section is given in Table 1.

**Table 1.** Overview of RL in medical image object and lesion detection.

| Author | ROI | Modality | Algorithm | State | Action Number | Reward Design |
|---|---|---|---|---|---|---|
| (Luo et al., 2019) | Heart | OCT | ADNet | Spatial-temporal locations correlation information | 9 | Change of intersection-over-union (IOU) index |
| (Maicas et al., 2017) | Breast | MR | DQN | Current bounding volume | 9 | Change of intersection-over-union (IOU) index |
| (Navarro et al., 2020) | Lung, Kidney, Liver, Spleen, Pancreas | CT | DQN | Voxel value of the current bounding box | 11 | Change of intersection-over-union (IOU) index |
| (Pesce et al., 2019) | Lung | Xray | REINFORCE | The part of the image observed by the glimpse | Number of image pixels | Correctness of the classification; Location of the glimpse |
| (Zhang et al., 2021) | Vertebral Body | MR | SAC | Combination of the image patch, feature map and region mask. | 4 | Change of attention-focusing accuracy |
| (Stember & Shalu, 2020) | Brain | MR | DQN | The gaze plot the agent locate | 3 | Whether moving towards the gaze plot that include the tumor |
| (Zheng et al., 2021) | Prostate | MR | DQN | Voxel values contained in the bounding box | 4 | Change of the Distance and IOU between the bounding box and the target |

* Indicate that the missing part is not clearly defined in the original paper

## 3.2. *Medical Image Segmentation*

### 3.2.1. *Overview of Works*

*Threshold Determination*

(Sahba et al., 2006) is the first attempt at using RL for medical image segmentation. The key idea is to formulate this segmentation task as a control task by a simple Q-learning agent that decides the optimal local thresholds and the post-processing parameters. The quality of the segmentation is considered when designing the state. The segmentation threshold and size of the structuring elements are changed by taking a series of actions. Though simple as this initial research, the segmentation quality was acceptable while significantly reducing the required human interaction compared to the mainstream methods like active contour at that time.

*Pre-locate the Segmentation Region*

Most supervised-learning-based catheter segmentation methods require a large amount of well-annotated data. (Yang et al., 2020) proposed a semi-supervised pipeline shown in Figure 11 that first uses a DQN agent to allocate the coarse location of the catheter and then conducts patch-based segmentation by Dual-UNet. The RL agent reduced the need for voxel-level annotation in the pre-allocation stage. The semi-supervised Dual-UNet exploited plenty of unlabeled images according to prediction hybrid constraints, thus improving the segmentation performance. The states are defined as the 3D observation patches, and the agent can update the states by moving the patch center point (x, y, z) along the x, y, and z-axis of the observation space. Like the landmark detection problems, the agent would give a negative reward if the patch moves away from the target; otherwise, a positive reward for moving toward and no reward if standing still. Compared to the state-of-the-art methods, this proposed pipeline requires much less computation time and achieves a minimum of 4% segmentation performance improvement measured by Dice Score.

*Hyperparameters optimization*

Instead of directly involved in the segmentation process, RL agents can also be applied to optimize the existing medical image segmentation pipelines (Bae et al., 2019; Qin et al., 2020; Yang et al., 2019). (Bae et al., 2019) used RL as the controller to automate the searching process of optimal neural architecture. The required search time and the computation power are significantly reduced by sharing the parameters while adopting a macro search strategy. Tested on the medical segmentation decathlon challenge, the authors assert that this optimized architecture outperformed the most advanced manually searched architectures.

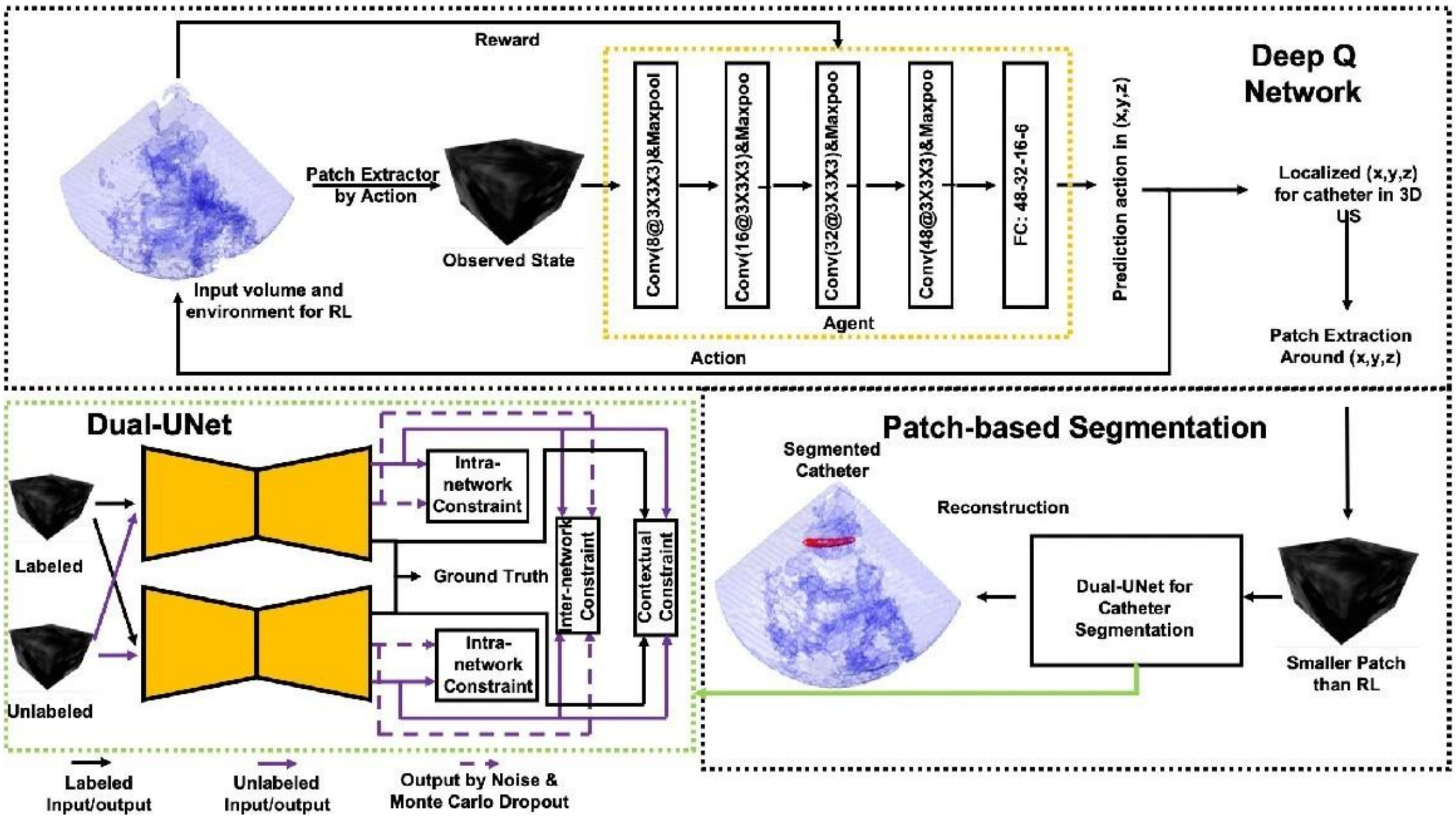

**Figure 11**: The semi-supervised DQN-driven catheter segmentation framework. Courtesy of (Yang et al., 2020).

Realizing the problem that some randomly augmented images might sometimes even harm the final segmentation performance, (Qin et al., 2020) implemented an automated end-to-end augmentation pipeline using Dual DQN (DDQN) agent. By making the trails and saving the experiences, the agent would learn to determine the augmentation operations beneficial to the segmentation performance according to the fed-back Dice ratio. Twelve different basic actions would change the state to achieve augmentation. The state is defined as the extracted feature from U-Net. It is interesting to observe that horizontal flipping and cropping are two of the most useful operation.

(Yang et al., 2019) from NVIDIA integrated the highlights of the previous two reviewed papers. With an RNN-based controller, this research automates the design process of hyper- parameters and image augmentation to explore the maximum potential of the state-of-the- art models. The optimal policy is learned using the proximal policy optimization to decide the training parameters. Tested on the medical decathlon challenge tasks, the RL searched model and augmentation parameters have shown remarkable effectiveness and efficiency.

*Segmentation as a Dynamic Process*

Observing that many existing automated segmentation pipelines may often fail in real clinical applications, (Liao et al., 2020) implemented multi-agent reinforcement learning to interact with the users that can achieve an iteratively refined segmentation performance. This multi- agent strategy captures the dependence of the refinement steps and emphasizes the uncertainty of binary segmentation results in states definition. Instead of defining the state as the binary segmentation result, it is formulated as $s_i^{(t)} = \left[b_i, p_i^{(t)}, h_{+,i}^{(t)}, h_{-,i}^{(t)}\right]$, where $b$, $p$, $h$ are the value, previous prediction probability, hint maps of voxel $i$, and $t$ indicates the current step. The actions will change the segmentation probability by an amount $a \in A$, where $A$ is the action set. Furthermore, the voxel-wise reward is defined as $r_i^{(t)} = \chi_i^{(t-1)} - \chi_i^{(t)}$, where $\chi$ is the cross entropy between the label $y_i$ probability $p_i$, to refine the segmentation more efficiently. The refined final segmentation result outperformed Min- Cut (Boykov & Kolmogorov, 2004), DeepGeoS(R-Net) (Wang et al., 2018), and InterCNN (Bredell et al., 2018) on all the BRATS20015, MM- WHS, NCI-ICBI2013 datasets. Though published earlier than the (Liao et al., 2020) and adopted the older RL method to learn the policy, (Wang et al., 2013) incorporated not only the user's background knowledge but also their intentions. The proposed framework follows a "Show-Learn- Act" workflow, which reduces the required interactions while achieving context-specific and user-specific segmentation.

### *3.2.2. Assessment*

Tackling the image segmentation problems using RL agents provides us with an effective way to further optimize existing pipelines, overcome a limited number of training data, and interact with users to incorporate prior knowledge. Despite the novices of these methods, limitations still exist. The various definitions of states and actions may significantly influence the precision of the segmentation. In most works, the states are updated by a series of limited-number discrete actions to determine the final segmentation contours. Another problem is that the state design makes the agent only observe local or global information at a step. It would be interesting to see some methods in the future that can enable the agent to make these two pieces of information observable to the agent at the same time. A summary of the works we reviewed in this section is given in Table 2.

**Table 2.** Overview of RL in medical image segmentation.

| Author | ROI | Modality | Algorithm | State | Action Number | Reward Design |
|---|---|---|---|---|---|---|
| (Bae et al., 2019) | Brain, Heart, Prostate | MR | REINFORCE | Hyperparameters of searched architecture | * | Change of the Dice Score |
| (Liao et al., 2020) | Brain, Heart, Prostate | MR | A3C | Combination of the voxel value, previous segmentation probability and hint maps | 6 | Decreased amount of the cross entropy |
| (Qin et al., 2020) | Kidney | CT | DDQN | Extracted high-level feature | 12 | Change of the Dice ratio |
| (Sahba et al., 2006) | Prostate | Ultrasound | Q Learning | Segmented Objects | 6 | Change of segmentation quality |
| (Wang et al., 2013) | Ventricle | MR | Policy-based | Local image appearance, anatomical details | Continuous | Difference between the location given by model and user |
| (Yang et al., 2019) | Atrium, Lung, Pancreas, Spleen | CT, MR | PPO | Parameter values | * | Change of segmentation performance |
| (Yang et al., 2020) | Catheter | Ultrasound | DQN | 3D patches | 6 | Change of distance to the target |

* Indicate that the missing part is not clearly defined in the original paper

### 3.3. *Medical Image classification*

Classification is one of the most basic tasks in medical image analysis. Common medical image classification tasks include disease diagnosis and prognosis, anomaly detection, and survivorship prediction. According to the extracted information from the image, a label among the predefined classes would be assigned. Since most of the classification methods are fully supervised, these methods may often fail in real clinical settings due to a lack of high-quality labeled data.

#### 3.3.1. *Overview of Works*

*Train with Limited Annotation*

To overcome this limitation, Stember and Shalu published two papers (Stember & Shalu, 2021a, 2021b) which used a DQN agent in tandem with a TD model for accurate image classification with a minimal training set. The main workflow of these two papers is identical, except that the labels in the second paper were extracted from clinical reports using an SBERT (Reimers & Gurevych, 2019). By overlaying the images with red or green masks, they managed to formulate the classification problem as a behavioral problem shown in Figure 12. The states are defined as the original greyscale image overlaid in green or red, where the red mask indicates a wrong prediction and the green mask indicates the correct prediction. The binary action (0 or 1) predicts the label of the image as normal (0) or tumor-containing (1). If the prediction is correct, a +1 reward would be given to the agent; otherwise, a -1 reward would be penalized. Compared with the supervised learning model trained on the same minimum dataset, the RL showed excellent overfitting resistance and high classification accuracy.

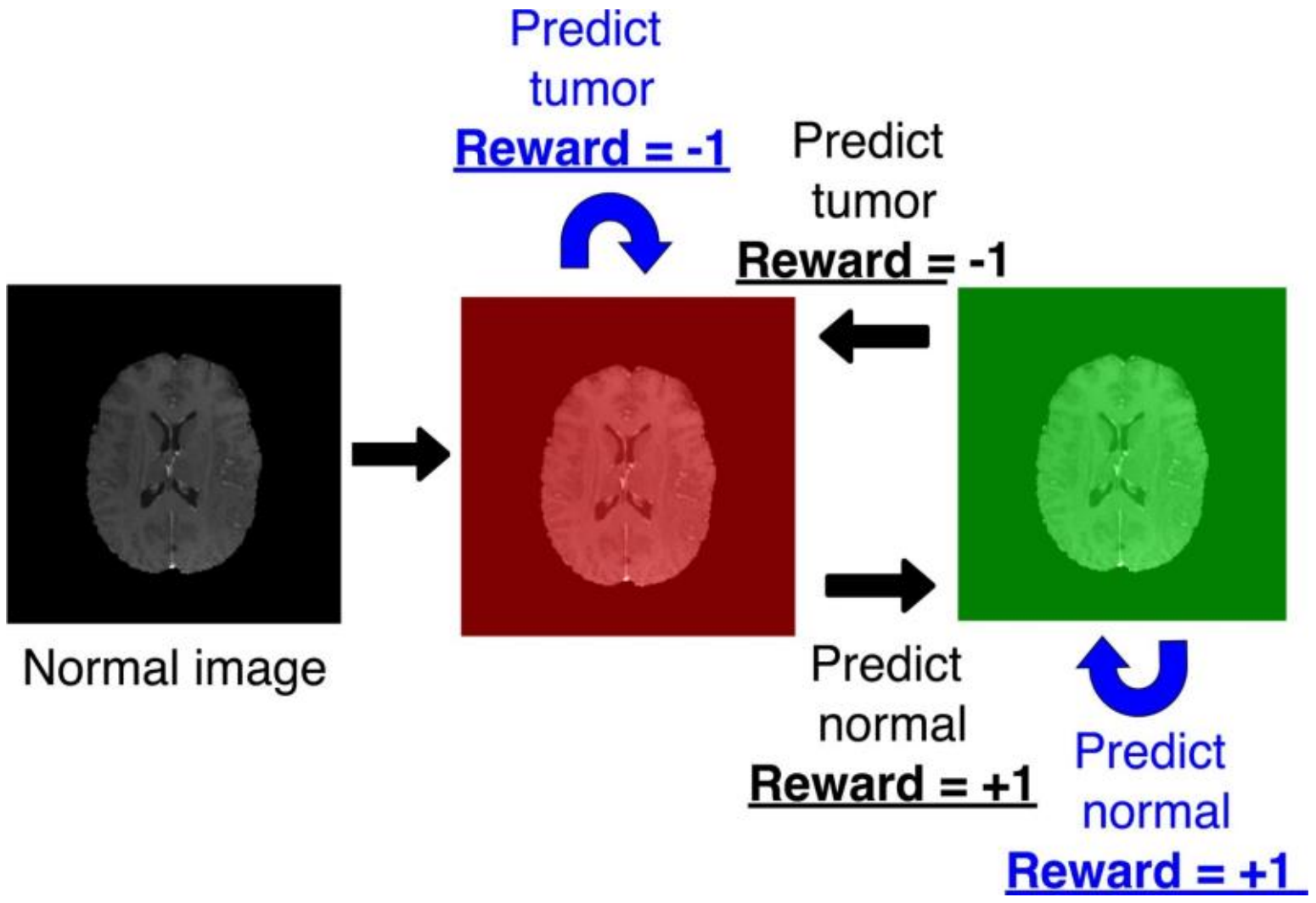

**Figure 12**: State transition and reward of a normal brain image. Courtesy of (Stember & Shalu, 2021a).

Another popular method to solve the lack of annotated data is to generate synthetic data. However, most medical synthesis pipelines do not assess the influence of the quality of these synthetic images in downstream tasks. Some misleading information in these synthetic data may skew the data distribution and thus harm the performance of the following tasks. To address this problem, (Ye et al., 2020) designed a PPO RL controller that can select synthetic images generated by HistoGAN. Considering the potential relationship between the generated and existing data, they used a transformer to output the action decision with the features extracted by a ResNet34 as input. The reward is designed according to the maximum validation accuracy in the last five epochs. Comparing the traditional augmentation, GAN augmentation, triplet loss metric learning, and centroid-distance-based selective augmentation, this Transformer-PPO-based RL selective augmentation achieved the best overall performance in the following classification task with an AUC score as high as 0.89.

*Optimal sample/weight/ROI selection*

As one of the common situations in the clinical setting, radiologists may have obtained considerable images, but the annotation process might be time-consuming and labor- expensive. A common approach to alleviate this problem is using active learning methods to select the most informative samples for annotations to improve the following tasks' performance. (Jingwen Wang et al., 2020), for the first time, formulated active learning for medical image classification as an automated dynamic Markov decision process. The current state consists of all the predicted values of the unlabeled images. This state then will be updated by continuous actions to decide the unlabeled ones for annotations according to the optimal policy. The model is trained according to the deep deterministic policy gradient algorithm (DDPG), which consists of an actor and a critic. A novice reward was designed to encourage the actor to focus on those samples that are incorrectly classified. Compared to other selective strategies, this RL framework achieved the best F1 score with all different percentages of training samples and a remarkable 0.70 score with only 40% labeled training data needed. (Smit et al., 2021) combined meta-learning and deep reinforcement learning for selective labeling. A bidirectional LSTM (BiLSTM) is used as the selector, and a non-parametric classifier is used as the predictor. Instead of using the RL agent as a controller, this work used the policy-gradient algorithm to optimize the objective function of the controller.

In clinical practice, an experienced physician usually makes diagnosis decisions from a combination of multi-modal images. Similarly, many state-of-the-art methods attempt to extract and integrate the information from various modalities to improve prediction performance. However, the weights of different modalities in this combination are hard to determine. (Jian Wang et al., 2020) automated this as an end-to-end process controlled by a REINFORCE RL agent, as shown in Figure 13. There are four US modalities involved in this pipeline: B-mode, Doppler, SE, and SWE. The model parameters are updated according to the global loss, which is a weighted summation of loss from each of the four modalities and a fusion loss calculated from the concatenated features. The states (weights) are updated (-0.2 or +0.2 or 0) at each step. Compared with other advanced single-modality and multi-modality methods on breast ultrasound datasets, this auto-weighted RL method achieved the best overall performance with an accuracy as high as 95.43%.

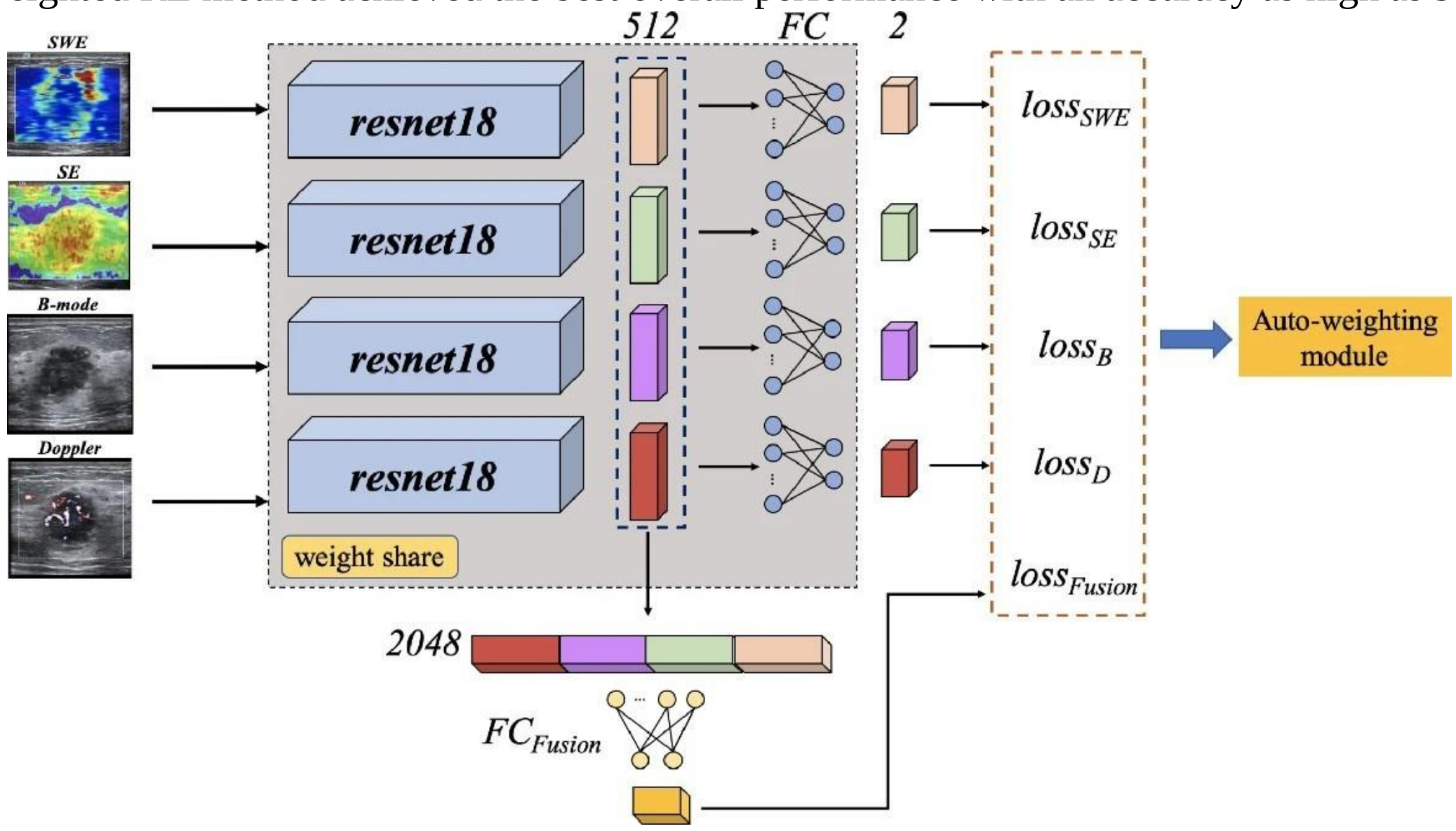

**Figure 13**: Schematic of the proposed multi-modal model. The weights of the losses are determined by the RL module. Courtesy of (Jian Wang et al., 2020).

For some types of medical images, histopathology images, for example, the resolution could be extremely high. Even though we have sufficient labeled data, it is still hard to perform the classification tasks due to the required high computation resources. Holding the belief that not all parts of the original images include valuable information, (Xu et al., 2020) proposed an RL-based pipeline for automated lesion region selection. Different from the hard-attention approaches, this RL selective attention method is end-to-end and fully automated, which includes two major stages - the selection stage and the classification stage. The backbone of the selection stage is a recurrent LSTM that outputs a binary action that decides whether a cropped patch is useful for classification. In the classification stage, a soft-attention network is used as the classifier. The reward that is fed back to the selector consists two-part, the training process reward, which represents the training stage accuracy, and the convergence reward, which represents the convergence performance. This selective strategy reduced the computation time by 50% and took less than 6ms to infer a single image.

*User Interaction*

Similar to the cases we have reviewed in the segmentation section, well-designed interactions with users can also help to improve the performance of classification tasks. However, instead of targeting the physicians as users, (Akrout et al., 2019) designed RL controlled QA system that interacts with the patients. A CNN is pre-trained to output the probabilities vector of the skin conditions. This vector is then concatenated with the vector of the patient's history answers about the symptom, where each binary value in the vector represents if a symptom is present. The action is designed to decide the next question to ask. The goal of the DQN agent is to maximize the possibility of classifying the correct condition when asking a specific question. Compared to CNN-only and decision-tree-based approaches, this RL-based symptom checker improved the classification accuracy up to 20% and 10%, respectively.

### *3.3.2. Assessment*

The nature of the image classification made it hard to define classification as a control problem. So instead of directly defining agent-based classification frameworks, most works attempted to use RL-agent to optimize the existing classification models' hyperparameters or the image preprocessing process. The only two papers that applied agents to the classification task itself came from the same authors, asserting that the agent-based classification method is superior to other methods on minimal training sets and still needs to be validated with more relevant studies. We look forward to seeing more research that can ingeniously design image classification as the control problem. A summary of the works we reviewed in this section is given in Table 3.

**Table 3.** Overview of RL in medical image classification.

| Author | ROI | Modality | Algorithm | State | Action Number | Reward Design |
|---|---|---|---|---|---|---|
| (Akrout et al., 2019) | Skin | Dermascope | DQN | Patient's history answers + output probability of pretrained CNN | 300 | Probability of correct condition if asked the question |
| (Smit et al., 2021) | Chest | Xray | Policy Gradient | * | * | * |
| (Stember & Shalu, 2021a) | Brain | MR | DQN + TD | Image overlaid in red or green | 2 | Classification correctness |
| (Stember & Shalu, 2021b) | Brain | MR | DQN + TD | Image overlaid in red or green | 2 | Classification correctness |
| (Jian Wang et al., 2020) | Breast | Ultrasound | REINFORCE | Weights | 3 | Classification correctness |
| (Jingwen Wang et al., 2020) | Chest | CT | DDPG | Predictions of the unlabeled images | Continuous | Possibility of being classified incorrectly |
| (Ye et al., 2020) | Cervix, Lymph Node | Histopathology | PPO | Selected images | 2 | Maximum validation accuracy of last epochs |
| (Xu et al., 2019) | Breast | Histopathology | Policy Gradient | Learning status representation + Incoming data statistics | 2 | Performance of the selection mechanism |

* Indicate that the missing part is not clearly defined in the original paper

### 3.4. *Medical Image Synthesis*

Medical image synthesis is the process of generating synthetic images that include artificial lesions or anatomical structures of a particular or multiple image modalities. There are typically two types of medical image synthesis (Wang et al., 2021): (i) inter-modality: transform the image of a specific modality to another modality, e.g., CT to MRI; (ii)intra- modality: transform the imaging protocol within the same modality, e.g., T1 sequence MRI to T2 sequence, or generating new images according to existing same-modality images. Nowadays, medical image synthesis algorithms are dominated by Generative Adversarial Networks (GANs) (Goodfellow et al., 2014), as their capability of providing visual appearing synthetic images with large diversity. Numerous GANs with different frameworks and techniques, such as Bayesian Conditional GAN (Zhao et al., 2020), progressive growing GAN (Guan et al., 2022), self-attention module (Lan et al., 2021), and deep supervision (Pan et al., 2021), have proposed and achieved state-of-the-art performance in various medical image synthesis applications. For the comprehensiveness of our review, we expand the concept of image synthesis a little while including the radiation dose map planning in this section.

#### 3.4.1. *Overview of Works*

*Semantic Map Generation*

(Krishna et al., 2021) successfully combined reinforcement learning and style transfer techniques to synthesize fine CT images from a small image dataset. There are two major steps in this pipeline. First, a DQN agent automatically generates the semantic maps of the lung CT images. Next, the B-splines and PCA interpolation are adopted to interpolate the semantic masks, thus providing texture information. The generated images have high resolution and are realistic enough to be used in other image analysis tasks.

*Pixel Value Alteration*

By designing a novice pixel-level graph RL method, (Xu et al., 2021) generates gadolinium-enhanced liver tumor images from non-enhanced images, thus avoiding the injection of toxic contrast agents. It is worth noting that this is the first paper that used RL agent for image synthesis and also the first attempt at designing agents, actions, and rewards all at the pixel level. The design of agents is based on the actor-critic structure but also integrates the idea of graph CNN. This graph-driven context- aware mechanism enables the model to capture both the small local object and global features. The reward function: $r_i^t = r(e)_i^t + \lambda(w)_i^t$ considered both the pixel-level (first item) and region-level (second item) rewards to make the measurement more accurate. According to the current states and rewards, the pixel-level agents determine the actions to increase, decrease or keep the pixel intensity. Trained and tested on 24375 images, the model outperformed the existing state-of-the-art method (Zhao et al., 2020).

*Synthetic Sample Selection*

Instead of focusing on directly synthesizing medical images, RL can also be applied in synthetic image assessment and selection. (Ye et al., 2020) selected HistoGAN-generated synthetic images according to the reliability and informativity. The selection process is formulated as a model-free, policy-based RL process stabilized by the Proximal Policy Optimization (PPO). The binary action is the output of the transformer model to decide if a fake image should be selected or discarded. The reward is designed to impel better accuracy in the following classification task. Compared with the unselected case, image augmentation using the chosen synthetic images improved the classification accuracy by 8.1% on the cervical data and 2.3% on the lymph node data.

RL has been widely applied in radiotherapy plan optimization to determine the optimal adaption time(Ebrahimi & Lim, 2021), tune the machine parameters (Hrinivich & Lee, 2020) and decide the beam orientation (Sadeghnejad-Barkousaraie et al., 2021). However, considering the scope of our review, we will only discuss agent-based dose map planning here.

*Dose Plan Generation*

(Shen et al., 2019) is among the first works that attempted to use the RL agent to optimize the dose-volume-histogram (DOV) in a step-by-step manner, leading to the final dose map. The states are defined as the current weights of the DOV, and five actions per weight will update the states. The reward is defined as the change in the dose plan's quality, according to the WTPN's guidance. Compared with humans, the RL agents led to an average improvement of 10.7% of the final dose plan quality in high dose-rate brachytherapy.

In one of their followed works (Shen et al., 2020), they applied the same idea to the external beam radiotherapy for prostate cancer patients. An end-to-end virtual treatment network (VTPN) was built to generate the optimum dose plan. They used ten patients for training and 64 patients for testing. With this VTPN-based treatment planning pipeline, the original ProKnown score increased to 10.93 from 6.07.

### 3.4.2. *Assessment*

Works that use RL agents for image synthesis are still scarce. There are three types of different strategies for the general meaning synthesis: semantic map generation, pixel-level value alteration, and synthetic sample selection. Compared to other state-of-the-art methods, the agent-based approach does not perform significantly superior and is training-time-consuming. According to our observation, the authors of these agent-based synthesis works did not publish any related results further, showing a fading away interest.

The key idea for the agent-based dose map generation is to update the DVHs step by step. Though this improved the dose plan quality, the reward function is not fully based on the clinical criteria, and the plan quality is only measured according to the DVH, which is only a simple representation. More works are encouraged to evaluate and improve this agent-based method for the more challenging treatment planning scenarios. A summary of the works we reviewed in this section is given in Table 4.

**Table 4.** Overview of RL in medical image synthesis.

| Author | ROI | Modality | Algorithm | State | Action Number | Reward Design |
|---|---|---|---|---|---|---|
| (Krishna et al., 2021) | Lung | CT | DQN | Control points coefficient sequences | * | Classification results of the pretrained classifier |
| (Xu et al., 2021) | Liver | MR | Pix-GRL (AC-based) | Pixel values of current image | 3 | Improvement of each pixel (pixel level); Improvement of pixels and surrounding pixels (region - level) |
| (Ye et al., 2020) | Cervix, Lymph node | Histopathology | PPO | Synthetic images | 2 | Validation classification accuracy |
| (Shen et al., 2019) | Cervix | Dose Volume Histogram (DVH) | Q-learning | Current DVH weights | 4 x 5 | Change of the plan quality |
| (Shen et al., 2020) | Prostate | Dose Volume Histogram (DVH) | Q-learning | Current DVH weights | 5 x 5 | Change of the plan quality |

* Indicate that the missing part is not clearly defined in the original paper

### 3.5. *Medical Image Registration*

Image registration is the process of transforming two different images into the same co- ordinates. In the medical imaging domain, the registration could be inter-patient, and intra- patient (but at different time points), inter-modality (e.g., MRI, CT, CBCT, Ultrasound), inter- dimensionality (2D-3D). The transformation models may also vary depending on the properties of the registration pair. It can generally be categorized into rigid, affine, and deformable transformation, where the rigid transformation is the simplest one to achieve using parametric models. Traditionally, the registration is completed via optimizing the similarity metrics. However, the high dimensionality of the medical images and the tissue's deformations and artifacts make this method unstable and sometimes even infeasible. The emergence of agent- based methods tackled the registration problem from a totally different angle—the formulation of the MDP process. Controlled by RL agents, the registration tasks achieve unprecedented robustness and precision.

#### 3.5.1. *Overview of Works*

*Rigid Registration*

The first attempt to use an agent-based method to solve the registration problems was by (Liao et al., 2017). Unlike the standard methods that focus on matching metrics optimization, this agent-based approach attempted to find the optimum sequence of actions that can align the images for registration. To train the agent with limited data, they first got synthetic data by dealigning the labeled training pairs. The registration process is done in a two-step hierarchical manner to ensure robustness and accuracy. A coarse registration is first done in a broader field of view (FOV) with lower resolution, followed by the fine alignment on the full resolution image. This proposed method outperformed the ITK registration method, Quasi-global search, and semantic registration methods on both the spine and heart 3D-3D registration to a large extent. It proved the possibility and superiority of agent-controlled registration.

Based on the idea of (Liao et al., 2017), (Miao & Liao, 2019) evaluate the performance of this pipeline for 2D-2D and 3D-3D intra-modality registration. They discussed the state, action space, and the design of rewards in detail and this chapter of the book can serve as a good tutorial. In another work (Miao et al., 2018), they attempted to use multi-agent attention mechanism to solve the 2D-3D registration for images with severe artifacts. Instead of choosing the commonly used CNNs, this work adopted dilated fully-CNN (FCN) as the backbone of the agent. This strategy reduced the registration problem's degrees of freedom (DoFs) from 8 to 4, significantly improving training efficiency.

Moreover, they implemented the auto attention mechanism to overcome the strong arti- facts in the 2D images. The so-called attention is achieved by using multiple local agents to learn and decide the actions, and only the agents whose confidence scores are above a certain threshold (0.67) will be selected. There are six pairs of actions (negative and positive) of the Special Euclidean Group SE(3). The framework was tested on both CBCT and more complex surgical data. In the more complex scenario, the multi-agent attention mechanism showed much better robustness (less performance degradation) than the corresponding single-agent system.

*Non-rigid Registration*

So far, what we have tackled with are all rigid problems. (Krebs et al., 2017) first attempted non-rigid registration with limited number of real inter-object pair. Both the inter-object and intra-object pairs were used for training. The intra-object pairs were generated as an augmentation step to compensate for the insufficient number of inter-object pairs. Considering that non-rigid registration has more degrees of freedom than the rigid transformation, the authors built statistical deformation models to serve as a low-dimensional representation of the problem. While minimizing the possible number of actions, the robustness of the registration is also guaranteed by the fuzzy action control. Experiments were conducted on both the 2-D and 3-D MR prostate images with a median Dice score of 0.88/0.76.

*Lookahead Inference*

To further improve the registration performance of rigid registration, in a series paper from (Hu et al., 2021; Sun et al., 2019), they incorporated the trained network with the lookahead inference. More specifically, the long-short-term-memory-machine (LSTM), specialized in tackling sequential data, is used to extract the spatial-temporal features. The fixed and moving images formed a 3D tensor to represent the state updated by transformations, including translations (+/- 1 pixel), rotations (+/- 1 degree), and scaling (+/- 0.05). In the testing phase, to make sure that the agent can reach the terminal state, they adopted a Monte Carlo rollout strategy to simulate different searching trajectories. The final transformation matrix is calculated as the weighted average of matrices from all the trajectories. Compared with other regression-based and agent-based methods, the addition of the lookahead and Monte Carlo rollout mechanism improved the robustness and accuracy of multimodal image registration tasks.

### *3.5.2. Assessment*

From the works we have reviewed above, it is not hard to see that agent-based registration methods have comparable or even better performance than the intensity/deep- similarity-based methods. However, most of the papers can only solve the rigid-registration problems. For the non-rigid transformer, however, the high-dimensional state-space and large number of DoFs may impede the agents from efficient convergence. So, researchers may have to transform the transformation space into a lower-dimensional space before applying the RL agents for registration. Another problem is that the agent may inhere the inefficiency from some similarity metrics used as the loss function during the training process, so novice loss functions should be designed for the RL frameworks. Last but not least, the methods that directly predict the transformation is still developing fast these years. Many state-of-the-art papers are emerging, while only a few papers are looking into agent-based registration, showing the low interest of researchers yet in this field. A summary of the works we reviewed in this section is given in Table 5.

**Table 5.** Overview of RL in medical image Registration.

| Author | ROI | Modality | Algorithm | State | Action Number | Reward Design |
|---|---|---|---|---|---|---|
| (Hu et al., 2021) | Nasopharynx | CT-MR | A3C | 3D tensor composed of the moving and the fixed image | 8 | Distance between the transformed landmark and the reference landmark |
| (Krebs et al., 2017) | Prostate | MR-MR | Q Learning | Spatial transformation parameters | 30 | The reduction of distance between the current parameters to the ground truth parameters |
| (Liao et al., 2017) | Spine, Cardiac | CT-CBCT | Q Learning | The rigid-body transformation matrix | 12 | The reduction of distance between the current transformation to the ground truth transformation |
| (Ma et al., 2017) | Chest, Abdomen | CT-Depth Image | Dueling DQN | 3D tensor including cropped image pair | 6 | Small constant during the exploration, and big constant at termination with sign determined by change of constant |
| (Luo et al., 2021) | Brain, Liver | MR-MR, CT-CT | SPAC | Pair of fixed images and moving image | * | The change of the dice score |
| (Miao et al., 2018) | Spine, Cardiac | CT-CBCT | Dilated FCN | The rigid-body transformation matrix | 12 | The reduction of distance between the current transformation to the ground truth transformation |
| (Miao & Liao, 2019) | Spine | CBCT-Xray | Q Learning | Transformations in SE (3) | 12 | Reduction of distance to the ground truth transformation |
| (Sun et al., 2018) | Nasopharynx | CT-MR | A3C | Concatenation of the fixed and moving image | 8 | Distance between the transformed landmark and the reference landmark |

* Indicate that the missing part is not clearly defined in the original paper

# 4. Discussion

Designing the medical image analysis as the RL problems is not an easy thing. Generally, the pipeline can be concluded in four steps: (i) thoroughly understand the environment; (ii) correctly choose an algorithm that would work for your problem; (iii) meticulously design the states, actions, and rewards; (iv) patiently train the framework to converge and validate the results. Though the RL for medical image analysis is still a new research field and the way to formulate the problems may vary from person to person, we can still conclude some interesting strategies from this works. While some challenges still exist, we are still optimistic about the perspective of this field.

*Single Agent vs. Multi Agent*

Multi-agent reinforcement (MARL) is one of the trends in recent years, aiming to improve the performance of single agent when facing large-scale and complicated environments. We have witnessed a few attempts to use multi-agent frameworks to solve some complex medical image analysis tasks. In these works, the agents collaborate and share their knowledge and experience to obtain the maximum mutual reward. The joint actions of all agents lead to the state's transition, and each agent's reward depends on the mutual strategy. It is worthy to note that the MARL may not always work. First of all, the reward mechanism of the MARL is more complicated than its single-agent counterpart. It is crucial to design a proper reward signal to improve the speed of learning and convergence. Besides, the enlarged (joint) state and action space may consume more computation power and decrease the efficiency of the RL framework.

*Multiscale Strategy*

One common strategy in RL for medical image detection and segmentation problems is the multiscale searching strategy. The agent will first perform the task at the coarsest level (usually by sampling the original image). An ROI is extracted and rescaled to a finer resolution, and the task is performed on this ROI again. This process is repeated iteratively until the finest resolution is reached. This multiscale, fine-to-coarse strategy is beneficial for training RL frameworks and saves computational resources and time.

*Model-Free vs. Model-Based*

The agent-based frameworks we have reviewed all belong to the model-free category. However, the model-based algorithm is another important subclass of RL. One possible reason why researchers did not attempt to use model-based RL is that it is hard to form the internal model of the environment considering the high dimensionality and large size of medical image data. However, the model-based models have their own advantage over the model-free ones, which is the higher sample efficiency. We are looking forward to future works on model-based RL for medical image tasks with small-scale labeled data.

## *4.1. Challenges*

Many challenges still impede medical imaging researchers from applying RL in their works.

The long training time and heavily consumed computational resources are something we can't ignore before starting RL-related research. Though RL has proved its efficiency in the inference phase of many tasks, learning from numerous trials and errors means it often takes at least a few days longer on some of the cutting-edge GPUs.

Besides, as we mentioned before, the design of the RL problems can be tricky. A slightly different design of the state, action, or reward may lead to a totally different performance (some models may fail to converge). The choice of hyperparameters of the RL frameworks also depends on the designers' experience with low explainability. Researchers may take days to experiment to find the parameters that work.

The low stability and reproducibility are other primary concerns (Khetarpal et al., 2018). Following the same workflows, some RL-agent may fail to work as well as described in the original paper. It is even harder to perform similarly when the input data source is changed. Even more challenging is that the works have been scarce, but most of the papers in this field will not make their code publicly available.

One last but not least, there have been some more accessible state-of-the-art methods to solve some types of medical imaging problems, and RL did not show a prominent advantage over them. For example, using GANs for medical image synthesis is still mainstream, and many outstanding papers are emerging each year, causing people's interest in using agent-based methods in these fields to fade away.

### *4.2. Future Perspectives*

The field of reinforcement learning developed very fast these years, and a lot of new theories or strategies have been proposed. However, the applications of RL in medical image analysis did not keep pace with these improvements. Here, we summarized some improvements that may lead to the future trend of agent-based medical imaging.

*Hierarchical Reinforcement Learning*

Hierarchical reinforcement learning (HRL) aims to improve the agent's efficiency facing some complicated problems. The main idea is to disassemble the final task into several subtasks in a hierarchical structure. There are three major subclasses of this type of framework: (i) HRL based on spatiotemporal abstraction and intrinsic motivation (Kulkarni, Narasimhan, et al., 2016); (ii) HRL based on internal option (Ravindran & IIL); (iii) Deep successor RL. These methods can potentially improve the current agent-based pipeline when solving some high-dimensional 3D image data or even 4D tracking data (Kulkarni, Saeedi, et al., 2016).

*Multitask and Transfer Reinforcement Learning*

For all the works reviewed, the trained RL agent can only perform the specific task trained for. While transfer learning is prevalent in deep learning for medical image analysis, it is reasonable to consider reusing some pre-trained RL agents for similar but different new tasks. Agent-based transfer learning can be categorized as behavioral transfer and knowledge transfer(Vithayathil Varghese & Mahmoud, 2020). By implementing the transfer learning in the current RL frameworks, we no longer have to train the agent to learn the complete policy from scratch. This will significantly reduce the training time and improve the frameworks' generalizability.

*Active Reinforcement Learning*

Active learning has two-fold meaning here: (i) Interaction with the users to incorporate users' prior knowledge; (ii) The agent will decide what data to be labeled and what data it will be trained. This active learning strategy can help the agent to understand the users' intention and get maximum performance with the minimum annotated data. However, this requires the involvement of human users (physicians in our case) in the process, so it might not be easy to be implemented in practice.

## 5. Conclusions

In this work, we have witnessed the success of some researchers' work that ingeniously turn the traditional image analysis tasks into RL-style behavioral or control problems. The basic concepts of reinforcement learning are first recapped, and a comprehensive analysis of applications of RL agents for different medical image analysis tasks was conducted in different sections. Under each section, the formulations of RL problems are discussed in detail from different angles. As the essential elements of the RL systems, the choice of algorithms, state, actions, and reward are highlighted in the table in Appendix A. These RL-based methods provide us a way to think of the problems and create new paradigms for solving current obstacles. We hope that readers can find commonalities from these works, further understand the principles of reinforcement learning, and try to apply reinforcement learning in their future research.

## 6. Disclosure

The authors are not aware of any affiliations, memberships, funding, or financial holds that might be perceived as affecting the objectivity of this review.